\pdfoutput=1

\documentclass[11pt]{article}

\usepackage[]{EACL2023}

\usepackage{times}
\usepackage{latexsym}

\usepackage[T1]{fontenc}

\usepackage[utf8]{inputenc}

\usepackage{microtype}

\usepackage{inconsolata}

\usepackage{multirow}
\usepackage{comment}
\usepackage{booktabs}
\usepackage{colortbl}
\usepackage{xcolor}
\usepackage{hhline}
\usepackage{amssymb}
\usepackage{pifont}
\usepackage{amsmath}
\usepackage{graphicx}
\usepackage{hyperref}

\newcommand{\cmark}{\ding{51}}%
\newcommand{\xmark}{\ding{55}}%
\newcommand{\secref}[1]{\S\ref{#1}}
\DeclareMathOperator*{\argmax}{arg\,max}

\newcommand{\wlaslalltr}{WLASL$^{\text{train}}_{\text{all}}$}
\newcommand{\wlasllabtr}{WLASL$^{\text{train}}_{\text{w}\mathcal{P}}$}

\newcommand{\wlasllabval}{WLASL$^{\text{val}}_{\text{w}\mathcal{P}}$}

\newcommand{\wlaslallte}{WLASL$^{\text{test}}_{\text{all}}$}
\newcommand{\wlasllabte}{WLASL$^{\text{test}}_{\text{w}\mathcal{P}}$}
\newcommand{\wlaslunlabte}{WLASL$^{\text{test}}_{\text{w/o}\mathcal{P}}$}

\definecolor{LightBlue}{rgb}{0.9, 1, 1}
\newcolumntype{i}{>{\columncolor{LightBlue}}r}
\definecolor{LightRed}{rgb}{1, 0.9, 0.9}
\newcolumntype{j}{>{\columncolor{LightRed}}r}
\definecolor{Gray}{gray}{0.9}
\newcolumntype{k}{>{\columncolor{Gray}}c}
\definecolor{Gray}{gray}{0.9}
\newcolumntype{o}{>{\columncolor{Gray}}r}

\title{Improving Sign Recognition with Phonology}

\author{Lee Kezar \\
    University of Southern California \\
    \texttt{lkezar@usc.edu} \\ \And
    Jesse Thomason \\
    University of Southern California \\
    \texttt{jessetho@usc.edu} \\ \AND
    Zed Sevcikova Sehyr \\
    San Diego State University\\
    \texttt{zsevcikova@sdsu.edu} \\
}

\begin{document}
\maketitle
\begin{abstract}
We use insights from research on American Sign Language (ASL) phonology to train models for isolated sign language recognition (ISLR), a step towards automatic sign language understanding. 
Our key insight is to explicitly recognize the role of phonology in sign production to achieve more accurate ISLR than existing work which does not consider sign language phonology.
We train ISLR models that take in pose estimations of a signer producing a single sign to predict not only the sign but additionally its phonological characteristics, such as the handshape. 
These auxiliary predictions lead to a nearly 9\% absolute gain in sign recognition accuracy on the WLASL benchmark, with consistent improvements in ISLR regardless of the underlying prediction model architecture.
This work has the potential to accelerate linguistic research in the domain of signed languages and reduce communication barriers between deaf and hearing people. 

\end{abstract}

\section{Introduction}
When learning to recognize sign language, there is evidence that people rely on breaking signs down into their constituent parts, such as the configuration and location of the hand \citep{klima79}. 
This process is also true of spoken language recognition, where recognizing sound patterns plays a crucial role in one's ability to recognize a word. 
Sometimes, one of these ``parts'' (phonemes) is the only distinguishing factor between two very different terms, as seen in the signs for \textsc{difference} (palms up) and \textsc{balance} (palms down) in Croatian Sign Language \citep{csl}. 
Thus, the ability to encode and recognize individual phonemes and the relationships among them is essential for sign recognition. 
As a first step in exploring the practicality of phoneme recognition, we ask: Can machine learning models for isolated sign language recognition (ISLR) benefit from the phonological structure of signs?

\begin{figure}[t]
    \centering
    \includegraphics[width=\columnwidth]{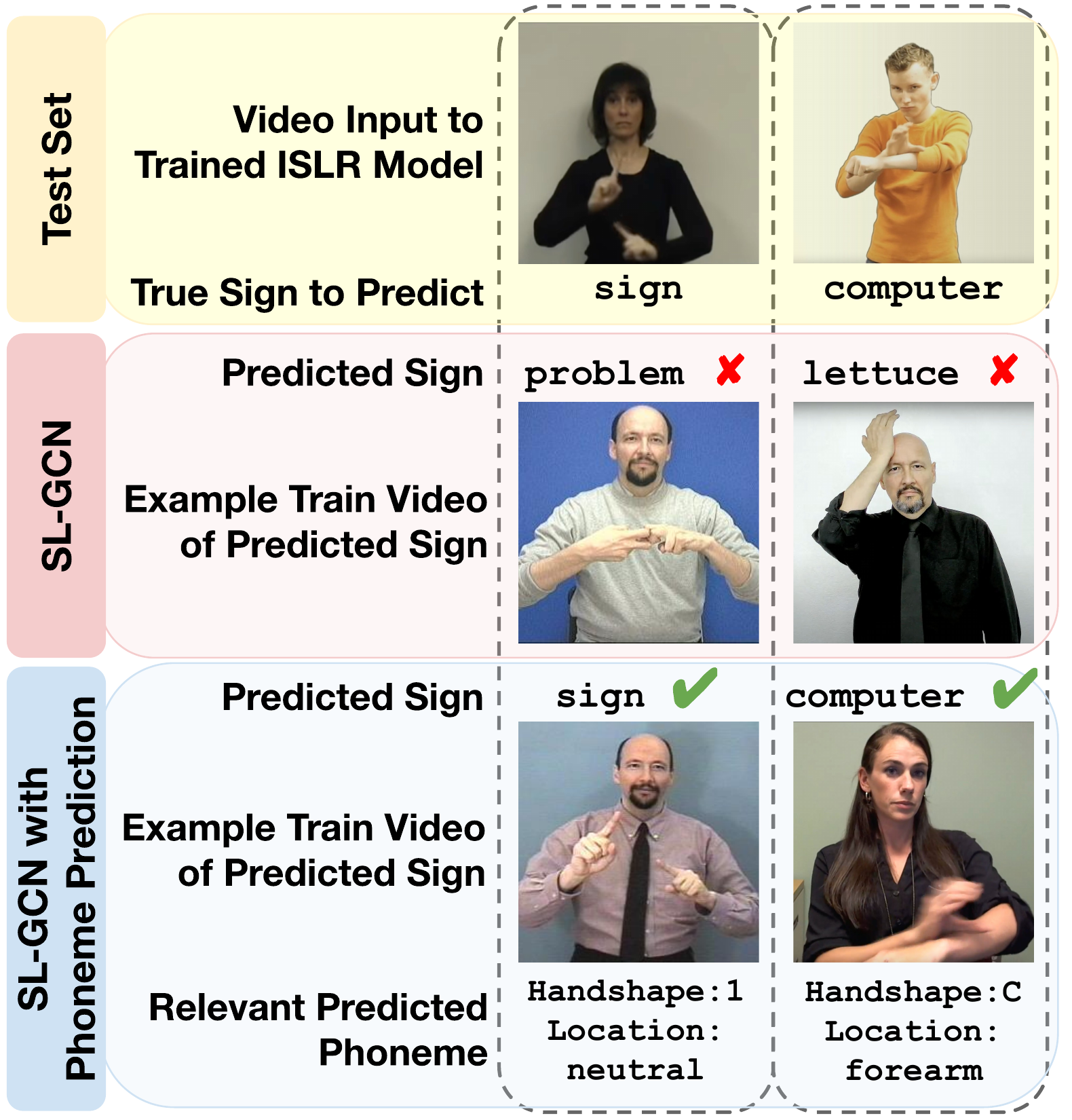}
    \caption{We demonstrate that sign language recognition models improve in accuracy when also tasked with predicting component phonemes of the sign.}
    \label{fig:teaser}
\end{figure}

While some ISLR models explicitly focus on the signers' hands \citep{handposeaware} or face \citep{bsl1k}, none have leveraged sign language phonology.
Instead, ISLR has been treated similarly to gesture recognition, where a ``gesture'' (such as swinging an imaginary bat or waving a hand) has no underlying structure except for that of the human body itself. 
This lack of structure might explain why state-of-the-art models like the Sign Language Graph Convolution Network (SL-GCN, \citealt{slgcn}) sometimes predict labels that are visually and phonologically unrelated to the ground truth, as shown in Figure~\ref{fig:teaser}.

In contrast, we show that models trained to recognize both signs and their phonemes will be more accurate at sign identification than those trained for ISLR alone. 
Our main contributions are:
\begin{itemize}
    \item We join an ISLR benchmark with a dataset of phonologically-labeled signs (\secref{method:data}) and describe a simple method for learning these labels alongside the target gloss\footnote{A ``gloss'' is a label for a sign that corresponds to its translation in the target language, such as \textsc{apple}.} (\secref{method:models}).
    \item We explore which and how many phoneme types are most beneficial as an auxiliary task to sign recognition (\secref{method:param_sel}, \secref{res:which_phons}).
    \item We demonstrate that adding auxiliary predictions for sign language phonology targets yields nearly 9\% absolute gain in accuracy for ISLR sign prediction (\secref{res:acc}), and that the resulting phoneme classification heads outperform prior work (\secref{res:phon_clf}).
\end{itemize}


\section{Background}
Sign languages are complete and natural languages primarily used by deaf and hard-of-hearing people. 
There are hundreds of sign languages in the world today collectively used by tens of millions of people \citep{ethnologue}. 
They rely on the hands, face, and body to communicate meaning according to complex grammars which are independent of any spoken language. 

Sign languages have been and continue to be largely overlooked in natural language processing (NLP) research, necessitating explicit calls for more inclusivity (e.g. \citealt{includingsl}, \citealt{slrgt}). 
In this paper, we seek to bridge robust techniques in NLP with insights from theories of sign language phonology.

\paragraph{Sign language phonology} is an abstract system of rules that governs how the structural units of signs (e.g., handshape, location, movement) are combined to create an infinite number of utterances. These manual units play a significant role at the phonological level similarly to place of articulation, manner, and voicing in spoken language. Theories of sign language phonology attempt to enumerate the meaningless units or ``phonemes'' found in a sign language and describe the complex relationships among them. 
In ASL-LEX 2.0, \citet{asllex} describe 16 types of phonemes, largely guided by Brentari’s Prosodic Model \citep{Brentari1998APM}. We provide three examples of these phoneme types here:

\begin{itemize}
    \item \textbf{Minor Location}: one of 37 regions of the body where the sign is produced (e.g. ``chin'').
    
    \item \textbf{Handshape}: one of 49 configurations of the hand (e.g. ``2'').
    
    \item \textbf{Path Movement}: one of 8 ways of moving the hand through space during the production of a single sign (e.g. ``circular'').
    
\end{itemize}


Brentari's Prosodic Model contains $<200$ possible phonemes across its 16 phoneme types, each of which can be observed during the production of any sign. 
In ASL-LEX 2.0, about 70\% of signs can be uniquely identified by their phonemes, making them an appealing conduit for learning to recognize signs. 
We leverage these properties by using them as target labels alongside the target gloss.

\paragraph{ISLR: Definition and Prior Work}
In ISLR, a model is given a video of one sign being produced in isolation and must predict the target gloss $S_{\textrm{gloss}}$.
Many models have been proposed to recognize isolated signs, varying with regard to input modality (e.g. pose, RGB video), pretraining (e.g. frame prediction, hand modeling), and encoding strategy (e.g. attention, convolution). 
\citet{openhands} provide a comprehensive framework for comparing models across multilingual data, in particular LSTMs \citep{lstm}, Transformers \citep{devlin-etal-2019-bert}, Spatio-Temporal Graph Convolution Network (ST-GCN, \citealt{stgcn}), and Sign Language Graph Convolutional Network (SL-GCN, \citealt{slgcn}). 

We evaluate our method with SL-GCN and via a Transformer network.
These models are open-sourced,\footnote{\url{https://openhands.readthedocs.io/}} easily modifiable, and take in pose information as input.
These models perform well on the WLASL 2000 benchmark \citep{wlasl}. 
While the model from \citep{handposeaware} obtains higher accuracy on that benchmark, their code is not publicly available to replicate those findings.

\section{Method}

We combine two datasets for the task of ISLR, ASL-LEX 2.0 \citep{asllex} and WLASL 2000 \citep{wlasl}, in order to learn ASL phonology (\secref{method:data}). 
Then, we describe how to utilize these data by learning two ISLR models to predict both the target gloss and the phonemes for any input (\secref{method:models}). 
Finally, we address the questions of how many and which phoneme types are best for ISLR (\secref{method:param_sel}). 
The dataset and modified models are released for replication and future work.\footnote{\url{https://github.com/leekezar/ImprovingSignRecognitionWithPhonology}}

\subsection{Data} \label{method:data}

We combine the phonological annotations in ASL-LEX 2.0 \citep{asllex} with signs in the WLASL 2000 ISLR benchmark \citep{wlasl}.
ASL-LEX contains 2,723 videos, each demonstrating a unique sign and human-annotated with phonemes across 16 categories. 
WLASL contains 21,083 videos, each demonstrating one of 2,000 unique signs (an average of 10.5 videos per sign). 

To combine these datasets, we edit the WLASL metadata file to add 16 new properties (one for each phoneme type) to each video example. 
If the video's English gloss is also found in ASL-LEX, then we copy the phonemes directly from ASL-LEX. 
If it is not found, then we set these new properties to \texttt{-1} and ignore them during training. 
After combining, 48\% of videos in the aggregated dataset have phonological labels, and all of the videos retain their original split (train, validation, and test) and English gloss.
Note that this dataset is identical in structure to WLASL-LEX \citep{wlasl-lex}, however, both our sources are more recently updated and contain more samples.
Table~\ref{tab:data_folds} provides a summary of the combined data.

\begin{table}[h!]
    \small
    \setlength{\aboverulesep}{0pt}
    \setlength{\belowrulesep}{0pt}
    \centering
    \begin{tabular}{krrrrr}
        \rowcolor[HTML]{CBCEFB} & & \multicolumn{4}{c}{\bf \# Videos} \\
        \rowcolor[HTML]{CBCEFB} \multirow{-2}{*}{\bf $\mathcal{P}$ Labels} & \multirow{-2}{*}{\bf \# Signs} & Train & Val & Test & \bf Total \\
        \toprule
        \xmark & 1246 & 7850 & 2221 & 1574 & 11645 \\
        \cmark & 754 & 6439 & 1695 & 1304 & 9438 \\
        \midrule
        \bf Total & 2000 & 14289 & 3916 & 2878 & 21083 \\
        \bottomrule
    \end{tabular}
    \caption{We match phonological data from ASL-LEX 2.0 with signs in the WLASL benchmark to create a subset of WLASL with phoneme type labels $\mathcal{P}$.}
    \label{tab:data_folds}
\end{table}

\begin{table*}[t!]
    \setlength{\aboverulesep}{0pt}
    \setlength{\belowrulesep}{0pt}
    \setlength{\tabcolsep}{2pt}
    \centering
    \begin{tabular}{kkrrriiijjj}
        \rowcolor[HTML]{CBCEFB}
         & Pred & \multicolumn{3}{c}{\wlaslallte} & \multicolumn{3}{c}{\wlasllabte} &
         \multicolumn{3}{c}{\wlaslunlabte}\\ 
        \rowcolor[HTML]{CBCEFB} \multirow{-2}{*}{Model} & $\mathcal{P}$ & \%A@1 & \%A@3 & MRR & \%A@1 & \%A@3 & MRR & \%A@1 & \%A@3 & MRR \\
        \toprule
		 & \xmark & $29.4$\tiny{$\pm1.6$} & $50.2$\tiny{$\pm2.3$} & $.43$\tiny{$\pm.02$} & $35.0$\tiny{$\pm1.8$} & $56.1$\tiny{$\pm1.7$} & $.48$\tiny{$\pm.02$} & $24.8$\tiny{$\pm1.4$} & $45.3$\tiny{$\pm3.0$} & $.39$\tiny{$\pm.02$} \\
		\multirow{-2}{*}{SL-GCN} & \cmark & $38.1$\tiny{$\pm0.5$} & $61.0$\tiny{$\pm0.3$} & $.52$\tiny{$\pm.00$} & $44.1$\tiny{$\pm1.1$} & $64.1$\tiny{$\pm0.6$} & $.56$\tiny{$\pm.01$} & $33.1$\tiny{$\pm0.3$} & $58.4$\tiny{$\pm0.2$} & $.49$\tiny{$\pm.00$} \\
            \cline{3-11}
		\multicolumn{2}{c}{\cellcolor{Gray}\emph{$\Delta$ Improvement}} & $^*8.7$\tiny{\phantom{$\pm0.0$}} & $^*10.8$\tiny{\phantom{$\pm0.0$}} & $^*.09$\tiny{\phantom{$\pm.00$}} & $9.1$\tiny{\phantom{$\pm0.0$}} & $8.1$\tiny{\phantom{$\pm0.0$}} & $.08$\tiny{\phantom{$\pm.00$}} & $8.3$\tiny{\phantom{$\pm0.0$}} & $13.1$\tiny{\phantom{$\pm0.0$}} & $.10$\tiny{\phantom{$\pm.00$}} \\
		\midrule
		 & \xmark & $20.5$\tiny{$\pm0.4$} & $36.9$\tiny{$\pm1.0$} & $.32$\tiny{$\pm.01$} & $24.5$\tiny{$\pm1.1$} & $41.2$\tiny{$\pm1.7$} & $.36$\tiny{$\pm.01$} & $17.2$\tiny{$\pm0.3$} & $33.3$\tiny{$\pm0.7$} & $.29$\tiny{$\pm.00$} \\
		\multirow{-2}{*}{Transformer} & \cmark & $23.4$\tiny{$\pm0.4$} & $41.7$\tiny{$\pm0.7$} & $.36$\tiny{$\pm.01$} & $28.2$\tiny{$\pm0.4$} & $46.5$\tiny{$\pm0.6$} & $.40$\tiny{$\pm.01$} & $19.3$\tiny{$\pm1.0$} & $37.8$\tiny{$\pm1.6$} & $.32$\tiny{$\pm.01$} \\
            \cline{3-11}
		\multicolumn{2}{c}{\cellcolor{Gray}\emph{$\Delta$ Improvement}} & $^*2.8$\tiny{\phantom{$\pm0.0$}} & $^*4.8$\tiny{\phantom{$\pm0.0$}} & $^*.04$\tiny{\phantom{$\pm.00$}} & $3.7$\tiny{\phantom{$\pm0.0$}} & $5.3$\tiny{\phantom{$\pm0.0$}} & $.04$\tiny{\phantom{$\pm.00$}} & $2.1$\tiny{\phantom{$\pm0.0$}} & $4.5$\tiny{\phantom{$\pm0.0$}} & $.03$\tiny{\phantom{$\pm.00$}} \\
		\bottomrule
    \end{tabular}
    \caption{ISLR model performance with and without training with auxiliary phoneme predictions averaged over four seeds.
    Models are trained on \wlaslalltr\ and evaluated on \wlaslallte.
    Models trained to predict phonemes improve over their ISLR-only baselines on \textit{both} signs seen at training time with phonemes (\wlasllabte) and signs for which no phonological data was available during training (\wlaslunlabte).
    Differences on \wlaslallte\ are significant ($^*$) at $p<0.05$ under a Welch's two-sided $t$-test with a Bonferroni correction applied.
    }
    \label{tab:all_results}
\end{table*}

\subsection{Models} \label{method:models}
We add phoneme value predictions to two ISLR model architectures: a graph convolutional network, SL-GCN~\cite{slgcn}, and a Transformer-based model. 
These models are implemented by the OpenHands project \citep{openhands} and are largely left untouched; we refer the reader to the OpenHands paper and code for implementation details.
The SL-GCN model treats pose estimations over time as a connected graph and learns 10 convolution layers over this graph, using spatial and temporal attention. 
The Transformer model treats pose estimations as a sequence of coordinates over time and learns 5 Transformer layers similarly to BERT~\cite{devlin-etal-2019-bert}. Importantly, both of these models implement spatial and temporal attention, a feature which enables phoneme recognition even when the phoneme exists for a short amount of time.

\begin{figure}[t]
    \centering
    \includegraphics[width=\columnwidth]{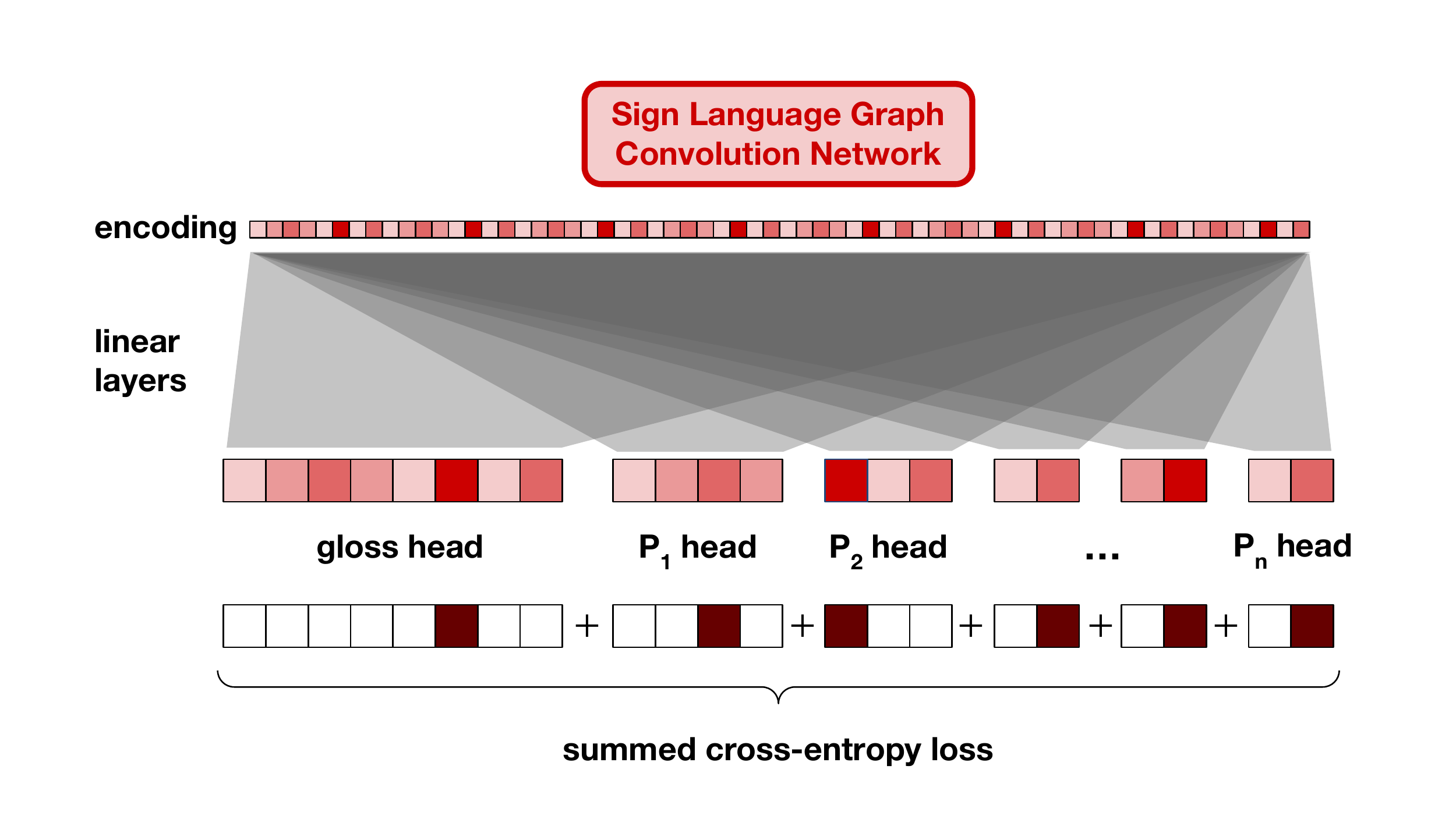}
    \caption{The proposed decoder relies on fully-connected layers for each classification head: one for the target gloss and one for each of the $n$ phoneme types.}
    \label{fig:model}
\end{figure}

For each model, we modify the decoder to classify not only the target gloss, but also the selected phonemes. 
This is accomplished by adding $n$ fully-connected layers to the decoder, each with shape \texttt{(hidden size, \# phoneme values)}. 
During the forward pass, the video encoding is used as the input for each fully connected layer (not chained together).
The total loss is then computed as:

$$ \mathcal{L}_{\textrm{total}} = \mathcal{L}_{\textrm{gloss}} + \sum_{\textrm{phoneme} \in \mathcal{P}} \mathcal{L}_{\textrm{phoneme}},$$

where $\mathcal{L}_{\textrm{gloss}}$ is the cross entropy of the model's gloss predictions, while $\mathcal{L}_{\textrm{phoneme}}$ is the cross entropy of the model's phoneme predictions. 
The sum of these losses is then backpropagated to the entire model, encouraging the encoder to learn a representation which more explicitly captures the desired phonemes $\mathcal{P}$ alongside the target gloss (Fig~\ref{fig:model}).

We train models until the validation accuracy has not improved in the last 30 epochs and use the top performing model for testing. For further details on model implementation and training procedure, see \citet{openhands}.

\subsection{Phoneme Type Selection} \label{method:param_sel}

It is not immediately clear which, if any, of the 16 phoneme types in ASL-LEX 2.0 yield improvements on ISLR.
Regarding \textit{how many} phoneme types, one might assume that more informative outputs would only improve a model's ability to recognize signs, and therefore all 16 phoneme types should be included.
However, these additions come at a cost to the encoder, which must now learn to fit more information into the same encoding space without adding new samples.
Furthermore, it is unclear \textit{which} types to maximize performance.

To address these questions, we define the utility $U$ of a set of phonemes types $\mathcal{P}$ as the percentage of signs that are uniquely identified by those types. 
Defined in this way, a set of phoneme types with high utility ensures that when a model can accurately predict those types, it is guaranteed to have sufficient information to recognize $U(\mathcal{P})$ percent of signs. 
$U(\mathcal{P})$ is provided by:

$$ U(\mathcal{P}) = \frac{\sum_{S, S' \in V} \mathbf{1}\left[p(S|\mathcal{P}) > p(S'|\mathcal{P})\right]}{|V|-1},$$

where $V$ is the set of all target glosses and $P(S|\mathcal{P})$ is the probability of a sign $S$ given the observed phoneme values in $\mathcal{P}$.
We implement $P(S|\mathcal{P})$ with a simple look-up table for all possible combinations of the 16 phoneme types.
With this utility function in hand, we can define the optimal subset of $n$ phoneme types as:

$$\mathcal{P}^*(n) = \left\{ \argmax_{\mathcal{P}} U(\mathcal{P}) : |\mathcal{P}|=n \right\}.$$

\section{Results}

\label{results}
We demonstrate across-the-board improvements on ISLR when predicting phonemes alongside glosses.
We measure model performance via ISLR accuracy, both top-1 and top-3, as well as mean reciprocal rank (MRR), which ranges from 1 (correct sign given highest prediction score) to 1/2000 (correct sign given lowest prediction score).

\subsection{Not All Phonemes are Helpful.} \label{res:which_phons}
First, we explore which subsets of the 16 labeled phonemes are most beneficial for downstream ISLR. 
We train models with auxiliary losses for $\mathcal{P}^*(n), n \in \left\{ 2, 5, 9, 16 \right\}$ and report their performance in Table \ref{tab:nparams}.
With two classification heads---handshape and minor location---we most improve ISLR on \wlasllabval.

\begin{table*}[t!]
    \setlength{\aboverulesep}{0pt}
    \setlength{\belowrulesep}{0pt}
    \centering
    \begin{tabular}{olrrr}
        \rowcolor[HTML]{CBCEFB}  $n$ & $\mathcal{P}^*(n)$ & \%A@1 & \%A@3 & MRR \\
        \toprule
        0 & $\emptyset$ & 50.2 & 69.3 & .62 \\
        2 & Dominant Handshape, Minor Location & 55.7 & 74.7 & .67 \\
        5 & $\mathcal{P}^*(2)$ + Nondominant HS, Path Movement, Repeated Movement &51.5 & 69.3 & .62 \\
        9 & $\mathcal{P}^*(5)$ + 2nd Minor Loc., 2nd Handshape, Wrist Twist, Contact &52.5 & 69.3 & .64 \\
        16 & $\mathcal{P}^*(9)$ + Remaining 7 phoneme types & 54.3 & 72.7 & .65 \\
        \bottomrule
    \end{tabular}
    \caption{SL-GCN sign recognition accuracy when trained on \wlasllabtr\ with auxiliary predictions of the top-$n$ phoneme types $\mathcal{P}$ and tested on \wlasllabval. See \secref{method:param_sel} for the details of $\mathcal{P}^*(n)$.}
    \label{tab:nparams}
\end{table*}

\subsection{Predicting Phonemes Improves ISLR.} \label{res:acc}
Table \ref{tab:all_results} demonstrates that adding classification heads for handshape and minor location yield a 3--9\% gain on top-1 accuracy, 5--11\% gain on top-3 accuracy, and .04--.09 gain on MRR.
These gains are greater for signs trained with phonological labels, but extend to signs that do not have phonological labels as well!

\subsection{SL-GCN Performs Accurate Phoneme Classification.} \label{res:phon_clf}
\begin{figure}[ht]
    \centering
    \includegraphics[width=\columnwidth]{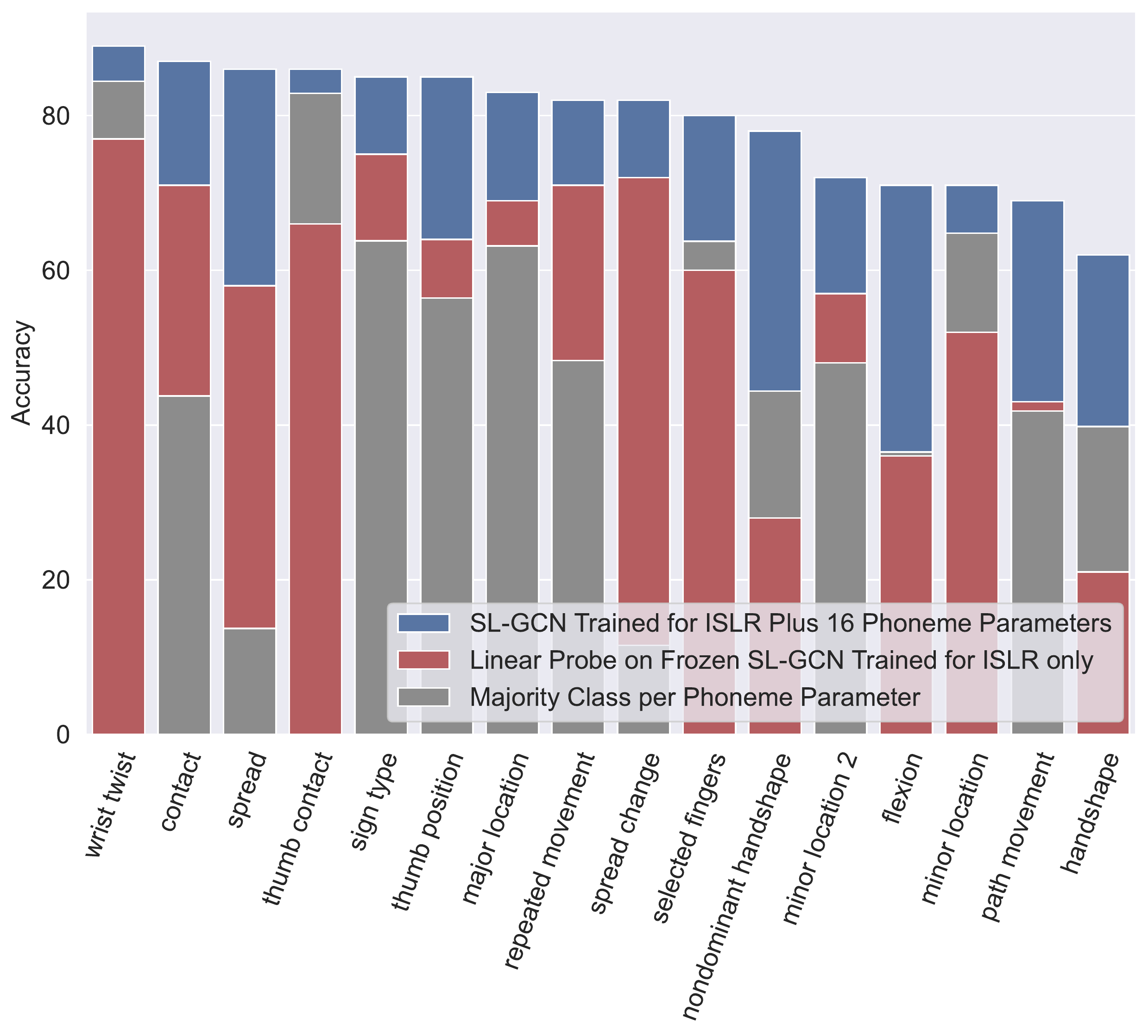}
    \caption{Model phoneme prediction accuracy when trained on \wlasllabtr\ and tested on \wlasllabte.
    }
    \label{fig:param_results}
\end{figure}

To lay the groundwork for modeling phonology in and of itself, we train SL-GCN to predict all 16 phoneme types and examine its accuracy at phoneme prediction (Figure~\ref{fig:param_results}).
We compare to a frozen SL-GCN encoder pretrained for \textit{only} ISLR, on top of which we learn linear probes for each phoneme type, as well as a majority class baseline.
In all cases, training SL-GCN explicitly for phoneme prediction leads to the highest phoneme prediction accuracy.
Prior work predicted phoneme values for Flexion, Major Location, Minor Location, Path Movement, Selected Fingers, and Sign Type~\cite{wlasl-lex}.
Despite not being the explicit goal of this work, SL-GCN with auxiliary phoneme prediction outperforms that model, too.

\section{Discussion}
We find that adding auxiliary classification tasks for sign phonemes to ISLR models statistically significantly improves sign recognition accuracy.
Representing phonemes during training may enable models to learn a more holistic latent representation of sign videos compared to models that only predict the target gloss.
The success of this approach provides evidence that handshape and minor location are not only useful in recognizing signs, but also easy enough to learn with semi-supervision (recall that only 48\% of the dataset has handshape and minor location labels). Our findings show that both models learn these new labels well (Fig~\ref{fig:param_results}) and as a result, the encodings for \textit{all} videos contain more relevant information for ISLR.

A secondary finding of this paper is that SL-GCN, when trained to recognize all 16 phoneme types, outperforms prior work by anywhere from 1\%-9\%. Still, there is room for improvement in phoneme recognition, especially for handshape, minor location, and path movement.

\section{Limitations} \label{disc:limitations}
The WLASL benchmark has several notable limitations that must be taken into account by those interested in using it. \citet{neidle} show that incorrect labels are pervasive in WLASL, causing lower ISLR accuracy and, in this work, incorrect phoneme labels. Additionally, existing sign language datasets do not provide information about the signers' fluency, dialect, age, or race and therefore may not be representative of those who use ASL. Finally, we caution those interested in collecting ASL data against scraping websites without permission, and we encourage acknowledging the creators of those sources. 

As a first attempt to model sign language phonology in order to improve sign recognition, we applied our approach to two models and used data for one language pair (ASL/English). Although many phonemes are shared across signed languages, more language pairs and models should be tested in order to verify our claim that learning phonology improves sign recognition \textit{in general}. In particular, the Two-Stream Inflated 3D ConvNet (I3D; \citealt{i3d}) model, designed for gesture recognition, has also been shown to do well on ISLR (e.g. \citealt{handposeguided}, \citealt{bsl1k}) and we look forward to extending our method to this model as well.

\bibliography{anthology,custom}

\begin{thebibliography}{19}
\expandafter\ifx\csname natexlab\endcsname\relax\def\natexlab#1{#1}\fi

\bibitem[{Albanie et~al.(2020)Albanie, Varol, Momeni, Afouras, Chung, Fox, and
  Zisserman}]{bsl1k}
Samuel Albanie, G{\"u}l Varol, Liliane Momeni, Triantafyllos Afouras, Joon~Son
  Chung, Neil Fox, and Andrew Zisserman. 2020.
\newblock {BSL-1K}: {S}caling up co-articulated sign language recognition using
  mouthing cues.
\newblock In \emph{European Conference on Computer Vision}.

\bibitem[{Bragg et~al.(2019)Bragg, Koller, Bellard, Berke, Boudreault,
  Braffort, Caselli, Huenerfauth, Kacorri, Verhoef, Vogler, and Morris}]{slrgt}
Danielle Bragg, Oscar Koller, Mary Bellard, Larwan Berke, Patrick Boudreault,
  Annelies Braffort, Naomi~K. Caselli, Matt Huenerfauth, Hernisa Kacorri, Tessa
  Verhoef, Christian Vogler, and Meredith~Ringel Morris. 2019.
\newblock Sign language recognition, generation, and translation: An
  interdisciplinary perspective.
\newblock \emph{The 21st International ACM SIGACCESS Conference on Computers
  and Accessibility}.

\bibitem[{Brentari(1998)}]{Brentari1998APM}
Diane Brentari. 1998.
\newblock \emph{A Prosodic Model of Sign Language Phonology}.
\newblock The MIT Press.

\bibitem[{Carreira and Zisserman(2017)}]{i3d}
Jo{\~a}o Carreira and Andrew Zisserman. 2017.
\newblock Quo vadis, action recognition? a new model and the kinetics dataset.
\newblock \emph{2017 IEEE Conference on Computer Vision and Pattern Recognition
  (CVPR)}.

\bibitem[{Cheng et~al.(2020)Cheng, Zhang, Cao, Shi, Cheng, and Lu}]{stgcn}
Ke~Cheng, Yifan Zhang, Congqi Cao, Lei Shi, Jian Cheng, and Hanqing Lu. 2020.
\newblock Decoupling gcn with dropgraph module for skeleton-based action
  recognition.
\newblock In \emph{European Conference on Computer Vision (ECCV)}.

\bibitem[{Dafnis et~al.(2022)Dafnis, Chroni, Neidle, and Metaxas}]{neidle}
Konstantinos~M. Dafnis, Evgenia Chroni, Carol Neidle, and Dimitri Metaxas.
  2022.
\newblock \href {https://aclanthology.org/2022.lrec-1.797} {Bidirectional
  skeleton-based isolated sign recognition using graph convolutional networks}.
\newblock In \emph{Proceedings of the Thirteenth Language Resources and
  Evaluation Conference}, pages 7328--7338. European Language Resources
  Association.

\bibitem[{Devlin et~al.(2019)Devlin, Chang, Lee, and
  Toutanova}]{devlin-etal-2019-bert}
Jacob Devlin, Ming-Wei Chang, Kenton Lee, and Kristina Toutanova. 2019.
\newblock \href {https://doi.org/10.18653/v1/N19-1423} {{BERT}: Pre-training of
  deep bidirectional transformers for language understanding}.
\newblock In \emph{Proceedings of the 2019 Conference of the North {A}merican
  Chapter of the Association for Computational Linguistics: Human Language
  Technologies, Volume 1 (Long and Short Papers)}, pages 4171--4186,
  Minneapolis, Minnesota. Association for Computational Linguistics.

\bibitem[{Eberhard et~al.(2022)Eberhard, Simons, and Fennig}]{ethnologue}
David~M. Eberhard, Gary~F. Simons, and Charles~D. Fennig. 2022.
\newblock \href {http://www.ethnologue.com} {Ethnologue: Languages of the
  world}.

\bibitem[{Hosain et~al.(2021)Hosain, Santhalingam, Pathak, Rangwala, and
  Kosecka}]{handposeguided}
Al~Amin Hosain, Panneer~Selvam Santhalingam, Parth~H. Pathak, Huzefa Rangwala,
  and Jana Kosecka. 2021.
\newblock Hand pose guided 3d pooling for word-level sign language recognition.
\newblock \emph{2021 IEEE Winter Conference on Applications of Computer Vision
  (WACV)}.

\bibitem[{Hu et~al.(2021)Hu, gang Zhou, and Li}]{handposeaware}
Hezhen Hu, Wen gang Zhou, and Houqiang Li. 2021.
\newblock Hand-model-aware sign language recognition.
\newblock In \emph{Association for the Advancement of Artificial Intelligence
  (AAAI)}.

\bibitem[{Jiang et~al.(2021)Jiang, Sun, Wang, Bai, Li, and Fu}]{slgcn}
Songyao Jiang, Bin Sun, Lichen Wang, Yue Bai, Kunpeng Li, and Yun~Raymond Fu.
  2021.
\newblock Skeleton aware multi-modal sign language recognition.
\newblock \emph{IEEE/CVF Conference on Computer Vision and Pattern Recognition
  Workshops (CVPRW)}.

\bibitem[{Klima and Bellugi(1979)}]{klima79}
E.S. Klima and U.~Bellugi. 1979.
\newblock \emph{The Signs of Language}.
\newblock Harvard University Press.

\bibitem[{Konstantinidis et~al.(2018)Konstantinidis, Dimitropoulos, and
  Daras}]{lstm}
Dimitrios Konstantinidis, Kosmas Dimitropoulos, and Petros Daras. 2018.
\newblock Sign language recognition based on hand and body skeletal data.
\newblock \emph{3DTV-Conference: The True Vision-Capture, Transmission and
  Display of 3D Video (3DTV-CON)}.

\bibitem[{Kuhn et~al.(2006)Kuhn, Ciciliani, and Wilbur}]{csl}
Ninoslava Kuhn, Tamara Ciciliani, and Ronnie Wilbur. 2006.
\newblock Phonological parameters in croatian sign language.
\newblock \emph{Sign Language \& Linguistics}.

\bibitem[{Li et~al.(2020)Li, Rodriguez, Yu, and Li}]{wlasl}
Dongxu Li, Cristian Rodriguez, Xin Yu, and Hongdong Li. 2020.
\newblock Word-level deep sign language recognition from video: A new
  large-scale dataset and methods comparison.
\newblock In \emph{The IEEE Winter Conference on Applications of Computer
  Vision (WACV)}.

\bibitem[{Sehyr et~al.(2021)Sehyr, Caselli, Cohen-Goldberg, and
  Emmorey}]{asllex}
Zed~Sevcikova Sehyr, Naomi~K. Caselli, Ariel Cohen-Goldberg, and Karen Emmorey.
  2021.
\newblock {The ASL-LEX 2.0 Project: A Database of Lexical and Phonological
  Properties for 2,723 Signs in American Sign Language}.
\newblock \emph{The Journal of Deaf Studies and Deaf Education}.

\bibitem[{Selvaraj et~al.(2022)Selvaraj, GokulN., Kumar, and
  Khapra}]{openhands}
Prem Selvaraj, C.~GokulN., Pratyush Kumar, and Mitesh~M. Khapra. 2022.
\newblock Openhands: Making sign language recognition accessible with
  pose-based pretrained models across languages.
\newblock In \emph{Association for Computational Linguistics (ACL)}.

\bibitem[{Tavella et~al.(2022)Tavella, Schlegel, Romeo, Galata, and
  Cangelosi}]{wlasl-lex}
Federico Tavella, Viktor Schlegel, Marta Romeo, Aphrodite Galata, and Angelo
  Cangelosi. 2022.
\newblock \href {https://aclanthology.org/2022.acl-short.49} {{WLASL}-{LEX}: a
  dataset for recognising phonological properties in {A}merican {S}ign
  {L}anguage}.
\newblock In \emph{Proceedings of the 60th Annual Meeting of the Association
  for Computational Linguistics (Volume 2: Short Papers)}.

\bibitem[{Yin et~al.(2021)Yin, Moryossef, Hochgesang, Goldberg, and
  Alikhani}]{includingsl}
Kayo Yin, Amit Moryossef, Julie~A. Hochgesang, Yoav Goldberg, and Malihe
  Alikhani. 2021.
\newblock Including signed languages in natural language processing.
\newblock In \emph{Association for Computational Linguistics (ACL)}.

\end{thebibliography}
\bibliographystyle{acl_natbib}

\clearpage

\end{document}